\documentclass{article}

\usepackage{arxiv}

\usepackage[utf8]{inputenc} 
\usepackage[T1]{fontenc}    
\usepackage{hyperref}       
\usepackage{url}            
\usepackage{booktabs}       
\usepackage{amsfonts}       
\usepackage{nicefrac}       
\usepackage{microtype}      
\usepackage{lipsum}
\usepackage{graphicx}
\graphicspath{ {./images/} }

\usepackage{algorithm}
\usepackage[]{algpseudocode}
\usepackage{sistyle}
\usepackage{makecell}
\usepackage{multirow}
\usepackage{wrapfig}
\usepackage{caption}
\usepackage{tabularx}
\usepackage{todo}
\usepackage{amsmath,amssymb}
\usepackage{xcolor}
\newcommand{\NEW}[1]{#1}

\title{Spatio-Temporal Outdoor Lighting Aggregation on Image Sequences using Transformer Networks}

\author{Haebom Lee$^{1,2}$ \and Christian Homeyer$^{1,3}$  \and Robert Herzog$^{1}$ \and Jan Rexilius$^{4}$ \and Carsten Rother$^{2}$}
\date{%
    $^1$ Corporate Research, Robert Bosch GmbH, Hildesheim, Germany\\
    $^2$ CVL Lab, Heidelberg University, Heidelberg, Germany\\
    $^3$ IPA Group, Heidelberg University, Heidelberg, Germany \\
    $^4$ Campus Minden, Bielefeld UoAS, Minden, Germany \\[2ex]%
    \today
}

\begin{document}
\maketitle
\begin{abstract}
In this work, we focus on outdoor lighting estimation by aggregating individual noisy estimates from images, exploiting the rich image information from wide-angle cameras and/or temporal image sequences. Photographs inherently encode information about the scene’s lighting in the form of shading and shadows. Recovering the lighting is an inverse rendering problem and as that ill-posed. 
Recent work based on deep neural networks has shown promising results for single image lighting estimation, but suffers from robustness. 
We tackle this problem by combining lighting estimates from several image views sampled in the angular and temporal domain of an image sequence. For this task, we introduce a transformer architecture that is trained in an end-2-end fashion without any statistical post-processing as required by previous work. Thereby, we propose a positional encoding that takes into account the camera calibration and ego-motion estimation to globally register the individual estimates when computing attention between visual words. We show that our method leads to improved lighting estimation while requiring less hyper-parameters compared to the state-of-the-art.
\end{abstract}



\section{Introduction}
Deep learning models are able to learn strong priors from data for solving highly ill-posed problems like single image reconstruction \cite{fan2017point}. In this manner, they have also been used for the task of lighting estimation.
The shading in a photograph captures the incident lighting (irradiance) on a surface point. It depends not only on the local surface geometry and material but also on the global (possibly occluded) lighting in a mostly unknown 3D scene. Different configurations of material, geometry, and lighting parameters may lead to the same pixel color, which creates an ill-posed optimization problem without additional constraints. Hence, blindly estimating the lighting conditions is notoriously difficult, and we restrict ourselves to outdoor scenes considering only environment lighting where the incident lighting is defined to be spatially invariant.


\begin{figure}[t]
	\begin{center}
		\centerline{\includegraphics[width=\textwidth]{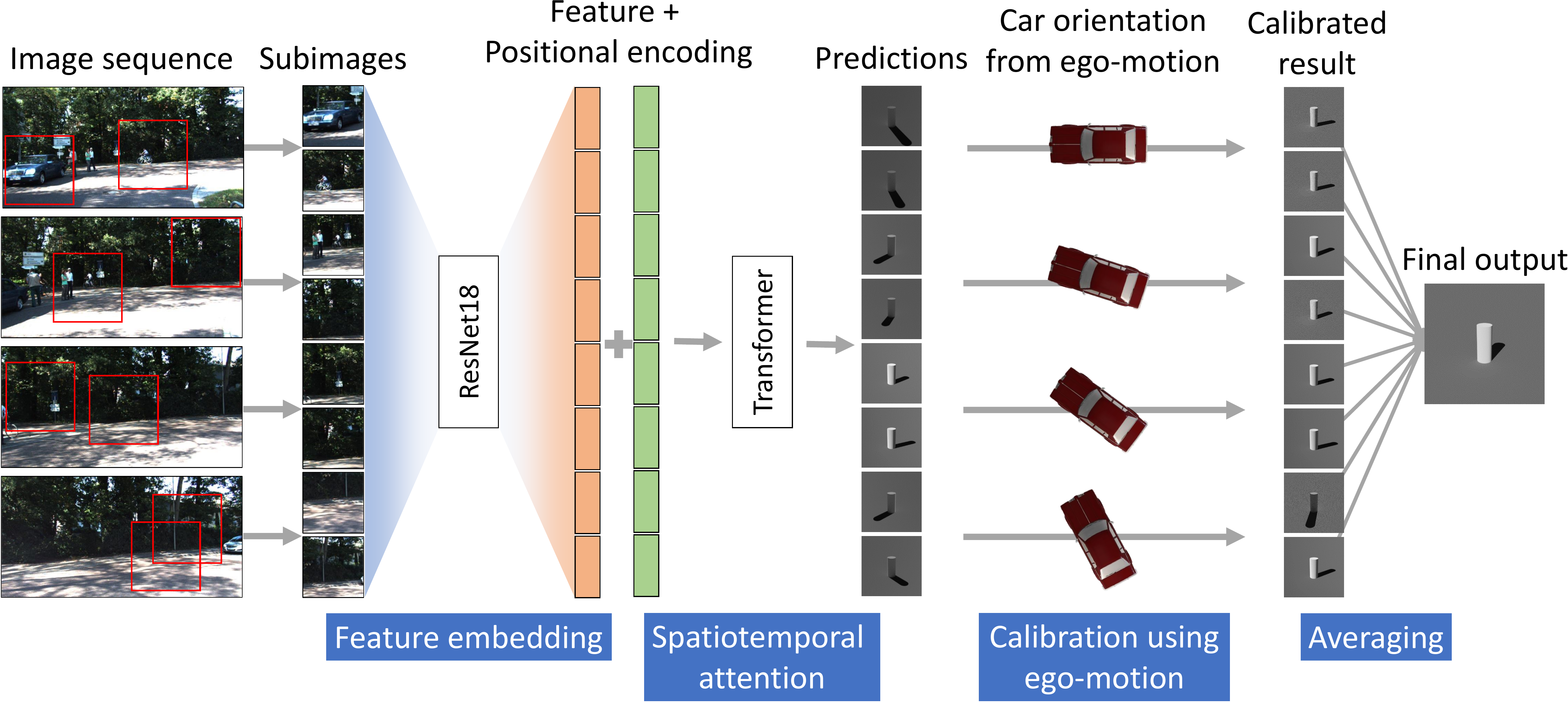}}
		\caption{\NEW{Spatio-temporal outdoor lighting aggregation on an image sequence: feature vectors are extracted from subimages using a pretrained ResNet18 network. Using an absolute positional encoding, our transformer network performs spatio-temporal attention. Individual estimates made in each camera coordinate system are calibrated using camera yaw angle data and fused to yield the lighting estimation for the sequence.}} 
		\label{fig:overview}
	\end{center}
\end{figure}

Estimating environment lighting can be regarded as the first step towards holistic scene understanding and enables several applications~\cite{balci2017sun,kan2019deeplight,madsen2011outdoor,wei2019simulating, zhu2021spatially}.
It is essential for augmented reality (seamlessly rendering virtual objects into real background images) because photo-realistically inserting virtual objects in real images requires knowing not just the 3D geometry and camera calibration, but also the lighting.
The human eye quickly perceives wrong lighting and shadows as unrealistic, and it has also been shown \cite{Dijk-Croon2019} that shadows are essential for single-image depth prediction using convolutional neural networks.


\NEW{Previous methods have focused on estimating explicit sky map textures~\cite{hold2019deep}, locating the sun position from single RGB images~\cite{hold2017deep,jin2020sun,zhang2019all}, calculating sun trajectories from longer time-lapse videos~\cite{balci2017sun,liu2012online} or in estimating a set of light sources in the context of RGBD-SLAM ~\cite{whelan2016elasticfusion}. 
In our work, we go in a similar direction as we robustly estimate the global sun direction by fusing estimates both from the spatial and temporal domain. 
The key is that we take advantage of known intrinsic calibration and ego-motion of multiple camera images, which all share the same direct sun light that is independent of relative translation. Therefore our method is applicable to both: pure rotational panorama images and images recorded with ego-motion as demonstrated in the results section.}


Image cues for resolving the lighting in a scene appear sparsely (e.g., shadows, highlights, etc.) or very subtle and noisy (e.g., color gradients, temperature, etc.). At the same time, not all images in a sequence provide the same quality of information for revealing the lighting parameters. For example, consider an image view completely covered in shadow.
Hence, the predictions for the lighting on individual images of a sequence are affected by a large amount of noise and many outliers.
To alleviate this issue we propose to sample many sub-views of an image sequence essentially sampling in the angular and temporal domain.
This approach has two advantages: First, we effectively filter noise and detect outliers, and second, our neural network-based lighting estimator becomes invariant to the imaging parameters like size, aspect ratio, and camera focal length and can explore details in the high-resolution image content.

\NEW{A preliminary version of this work has been published in \cite{lee2021spatiotemporal}. In this paper, we extend that work by using an end-2-end filtering approach that supersedes the statistical post-processing in \cite{lee2021spatiotemporal} by using a Transformer architecture \cite{dosovitskiy2020image,ranftl2021vision,girdhar2019video} which accounts for individual orientations and field-of-views of the input frames. In our experiments in Section \ref{sec:experiments}, we replace parts of our estimation pipeline and adapt the architecture of \cite{dosovitskiy2020image} for lighting source regression. To the best of our knowledge, we are the first to use an attention based model for the task of lighting estimation.}


We summarize our contributions as follows:
\begin{enumerate}
    \item Building on top of our preliminary work, we propose a spatio-temporal aggregation for sunlight estimation that is trained end-to-end using a \emph{Transformer} architecture.
    \item A novel handcrafted positional encoding tailored to the angular domain for sunlight estimation.
    \item More empirical results, showing the superiority of our new method.
\end{enumerate}




\section{Related Work}
\nocite{godard2019digging}
\nocite{panagopoulos2012simultaneous}
Outdoor lighting condition estimation has been studied in numerous ways because of its importance in computer graphics and computer vision applications~\cite{karsch2011rendering,lu2010foreground}. Related techniques can be categorized into two parts, one that analyzes a single image~\cite{hold2019deep,jin2019sun,lalonde2012estimating,ma2017find} and the other that utilizes a sequence of images \cite{balci2017sun,lalonde2014lighting,liu2012online,madsen2011outdoor}. For example, the outdoor illumination estimation method presented in \cite{madsen2005real} belongs to the latter as the authors estimated the sun trajectory and its varying intensity from a sequence of images. 
Under the assumption that a static 3D model of the scene is available, they designed a rendering equation-based~\cite{kajiya1986rendering} optimization problem to determine the continuous change of the lighting parameters. 
On the other hand, Hold-Geoffroy et al.~\cite{hold2017deep} proposed a method that estimates outdoor illumination from a single low dynamic range image using a convolutional neural network~\cite{krizhevsky2012imagenet} (CNN). The network was able to classify the sun location on 160 evenly distributed positions on the hemisphere and estimated other parameters such as sky turbidity, exposure, and camera parameters. 

Analyzing outdoor lighting conditions is further developed in \cite{zhang2019all} where they incorporated a more delicate illumination model~\cite{lalonde2014lighting}. The predicted parameters were numerically compared with the ground truth values and examined rather qualitatively by utilizing the render loss. Jin et al.~\cite{jin2020sun} and Zhang et al.~\cite{zhang2021outdoor} also proposed single image based lighting estimation methods. While their predecessors~\cite{hold2017deep,zhang2019all} generated a probability distribution of the sun position on the discretized hemisphere, the sun position parameters were directly regressed from their networks. Recently, Zhu et al.~\cite{zhu2021spatially} combined lighting estimation with intrinsic image decomposition. Although they achieved a noticeable outcome in the sun position estimation on a synthetic dataset, we were unable to compare it with ours due to the difference in the datasets.

The aforementioned lighting estimation techniques based on a single image often suffer from insufficient cues to determine the lighting condition, for example, when the given image is in complete shadow. Therefore, several attempts were made to increase the accuracy and robustness by taking the temporal domain into account~\cite{balci2017sun,lalonde2014lighting,madsen2011outdoor}. The method introduced in \cite{liu2012online} extracts a set of features from each image frame and utilizes it to estimate the relative changes of the lighting parameters in an image sequence. Their method is capable of handling a moving camera and generating temporally coherent augmentations. However, the estimation process utilized only two consecutive frames and assumed that the sun position is given in the form of GPS coordinates and timestamps~\cite{reda2004solar}. 

Lighting condition estimation is also crucial for augmented reality where virtual objects become more realistic when rendered into the background image using the correct lighting. Lu et al.~\cite{lu2010foreground}, for instance, estimated a directional light vector from shadow regions and the corresponding objects in the scene to achieve realistic occlusion with augmented objects. The performance of the estimation depends solely on the shadow region segmentation and finding related items. Therefore, the method may struggle if a shadow casting object is not visible in the image. Madsen and Lal~\cite{madsen2011outdoor} utilize a stereo camera to extend \cite{madsen2005real} further. Using the sun position calculated from GPS coordinates and timestamps, they estimated the variances of the sky and the sun over an image sequence. The estimation is then combined with a shadow detection algorithm to generate plausible augmented scenes with proper shading and shadows.

Recently, there have been several attempts utilizing auxiliary information to estimate the lighting condition~\cite{kan2019deeplight,xiong2021dsnet}. Such information may result in better performance but only with a trade-off in generality. K{\'a}n and Kaufmann~\cite{kan2019deeplight} proposed a single RGB-D image-based lighting estimation method for augmented reality applications. They utilized synthetically generated scenes for training a deep neural network, which outputs the dominant light source's angular coordinates in the scene. Outlier removal and temporal smoothing processes were applied to make the method temporally consistent. 
However, their technique was demonstrated only on fixed viewpoint scenes. Our method, on the other hand, improves its estimation by aggregating observations from different viewpoints. We illustrate the consistency gained from our novel design by augmenting virtual objects in consecutive frames.


\section{Proposed Method}
\label{sec:method}
We take advantage of different aspects of previous work and refine them into our integrated model. As illustrated in Fig.~\ref{fig:overview}, our model is composed of \NEW{two networks: a pre-trained ResNet18~\cite{he2016deep} and a transformer network~\cite{vaswani2017attention}.} We first randomly \NEW{crop several small subimages from a sequence of images}. Since modern cameras are capable of capturing fine details of a scene, we found that lighting condition estimation can be done on a small part of an image. \NEW{In this way, the samples obtained from each sequence provide different observations for the same global lighting condition.} 

\NEW{The ResNet18 network processes these images and yields patch embeddings. The input of the transformer network is then the sum of the patch embeddings and their corresponding cyclic 3D positional encodings (see Sec~\ref{sec:posenc}). The transformer network examines the noisy spatio-temporal observations and assigns proper attentions. The weighted features are delivered to a dense layer which outputs lighting condition estimates in the camera coordinate systems. Lastly, we perform a calibration step where we compensate the camera yaw angle of each subimage so that all estimates are in the unified global coordinate system. The final estimate of the given sequence is then the average of calibrated estimates. 
The assumption behind our spatio-temporal aggregation is that distant sun-environment lighting is invariant to the location the picture was taken and that the variation in lighting direction is negligible for short videos. Through the following sections, we introduce the details of our method.}



\subsection{Lighting Estimation}

There have been several sun and sky models to parameterize outdoor lighting conditions~\cite{hosek2012analytic,lalonde2014lighting}. \NEW{Although those methods are potentially useful to estimate complex lighting models, we focus only on the most critical lighting parameter: the sun direction. The rationale behind this is that ground-truth training data can easily be generated for video sequences having GPS and timestamp information (e.g., KITTI dataset~\cite{geiger2012we}). Therefore, the estimated lighting condition is given as a 3D vector $\vec v_{pred}$ pointing to the sun's location in the sequence.} 

Unlike our predecessors~\cite{hold2017deep,zhang2019all}, we design our network as a direct regression model to overcome the need for a sensitive discretization of the hemisphere.  
\NEW{The recent work of Jin et al.~\cite{jin2020sun} and Zhang et al.~\cite{zhang2021outdoor} presented regression networks} estimating the sun direction in spherical coordinates (altitude and azimuth). Our method, however, estimates the lighting direction using Cartesian coordinates and does not suffer from singularities in the spherical parametrization and the ambiguity that comes from the cyclic nature of the spherical coordinates.

Since we train our network in a supervised manner, \NEW{we compare the estimated sun direction with the ground truth and apply two more conditions to foster the training. The first loss function is defined to minimize the angle between the estimate and the ground truth sun direction} $\vec v_{gt}$:
\begin{equation}
\label{eq:l_light_cosine}
\begin{array}{l}
L_{cosine} = 1 - \vec v_{gt} \cdot \vec v_{pred} / \lVert\vec v_{pred}\rVert,
\end{array}
\end{equation}
with the two adjacent unit vectors having their inner product close to $1$. 
To avoid the uncertainty that comes from the vectors pointing the same direction with different lengths, we apply another constraint to the loss function:
\begin{equation}
\label{eq:l_light_norm}
\begin{array}{l}
L_{norm} = (1 - \lVert\vec v_{pred}\rVert)^2.
\end{array}
\end{equation}
The last term of the loss function ensures that the estimated sun direction resides in the upper hemisphere because we assume the sun is the primary light source in the given scene:
\begin{equation}
\label{eq:l_light_hemi}
\begin{array}{l}
L_{hemi} = max(0, -z_{pred}),
\end{array}
\end{equation}
where $z_{pred}$ is the third component of $\vec v_{pred}$, indicating the altitude of the sun. The final loss function is simply the sum of all terms as they share a similar range of values:
\begin{equation}
\label{eq:l_light}
\begin{array}{l}
L_{light} = L_{cosine} + L_{norm} + L_{hemi}.
\end{array}
\end{equation}



\subsection{Attention based Aggregation}
\label{sec:attention}

\NEW{In order to extract robust estimates from noisy observations, the aggregation process described in \cite{lee2021spatiotemporal} relies heavily on statistical filtering utilizing an outlier removal combined with the meanshift algorithm. 
However, this approach requires manual hyper-parameter tuning with handcrafted selection criteria. We extend this work by replacing the aggregation step with a purely end-to-end attention driven pipeline. The overview of our approach is illustrated in Figure~\ref{fig:overview}.}

\NEW{We take inspiration from \cite{dosovitskiy2020image} for our network design and adopt their hybrid architecture for our task. This includes self attention using multi-head attention layers \cite{vaswani2017attention} and preprocessing images with a pretrained convolutional neural network. Given a temporal sequence of $k$ images, we first select $n$ spatially random crops for each frame as done in our previous work~\cite{lee2021spatiotemporal}.  
On each crop, we apply a ResNet18~\cite{he2016deep} encoder to extract feature embeddings. 
Each embedded patch is fed as input to our transformer module for aggregation. The virtue of the transformer network is that it can associate observations from different space and time given a proper positional encoding.
Since all images patches share the same sun light and we assume we know their relative orientation due to the ego-motion estimation the Transformers attention mechanism inherently learns to filter the noisy patch-wise predictions. However, we need to provide the relative orientation of the patches in order to make the light estimation invariant to camera orientation, which we achieve via the \emph{positional encoding}. 
}


\subsection{Orientation-invariant Positional Encoding}
\label{sec:posenc}

Solely relying on image features enables only to estimate the lighting in the local camera frame. However, we need to fuse the estimates in a global reference frame in order relate different subimages. Since we assume sun-lighting, only the directional component of a recorded camera image is relevant to calibrate different frames. 
We inject this camera orientation in the image features via a positional encoding. However, we only encode the yaw angle of the camera rotations (the rotation around the ground-plane surface-normal) since pitch and roll angles are naturally captured in the image features of outdoor images (e.g, horizon). Further, we also encode the 2D position of the subimages cropped from the source frame independent of the intrinsic camera projection, i.e., in terms of viewing angles $\phi$ in the corresponding horizontal and vertical field of views. For example, the top left pixel gets a coordinate of $\left(-\frac{fov_h}{2},\; \frac{fov_v}{2}\right)$ for a pinhole camera model with a field of view of $fov_h$ and $fov_v$ horizontally and vertically respectively. 
To this end we concatenate the 2D angular image coordinate and the (temporal) camera rotation angle and apply a 3D cyclic positional encoding.
\NEW{We use an absolute positional encoding, i.e. 
\begin{align}
    x_i \longleftarrow x_i + p_i \; ,
\end{align}where the positional encoding $p_i$ and the subimage feature vector $x_i \in \mathbb{R}^d_x $ are superimposed. Similar to \cite{vaswani2017attention} we use a fixed encoding of sine and cosine functions with different frequencies. 
}

\begin{figure}[tb]
	\begin{center}
		\centerline{\includegraphics[width=\textwidth]{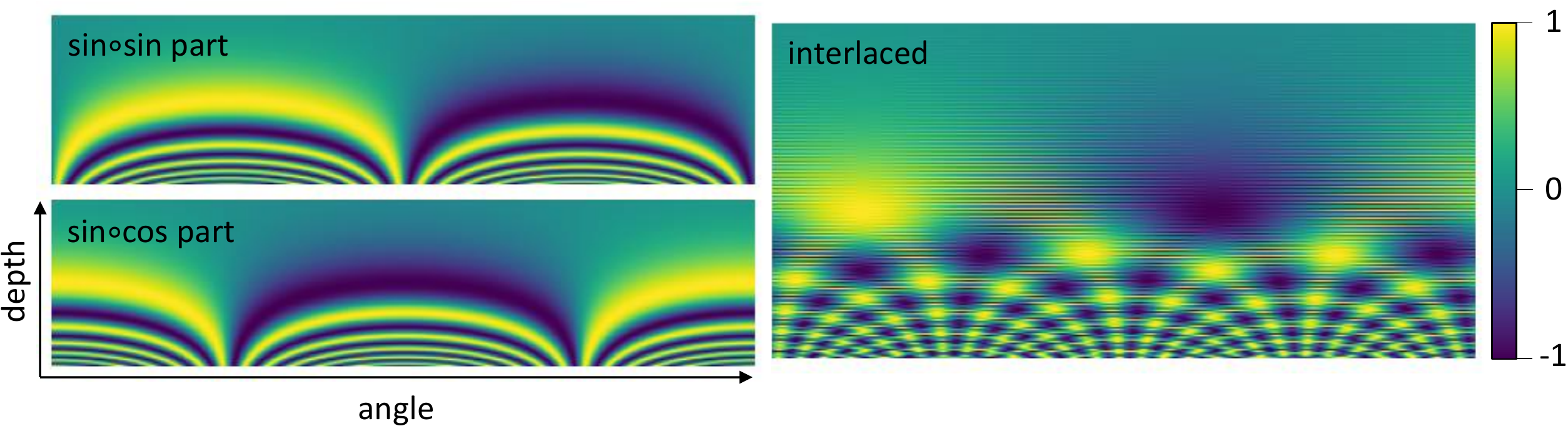}}
		\caption{\NEW{Cyclic positional encoding for angle $ \phi \in [0,\; 2\pi]$. 
		The periodicity of our encoding scheme is clearly visible on the left side images while their interlaced result on the right side shows its uniqueness for each angle.}}
		\label{fig:posenc}
	\end{center}
\end{figure}

\NEW{Since our positional encoding scheme encodes angles, it has to fulfill the following two conditions: 1) \textit{periodicity} - the transition from the encoding of $359^{\circ}$ to the encoding of $0^{\circ}$ should be as smooth as the transition from $0^{\circ}$ to $1^{\circ}$ and 2) \textit{uniqueness} - each angle should have a unique encoding. We present our cyclic positional encoding, satisfying those conditions, by using nested trigonometic functions as below:}

\begin{equation}
\label{eq:posenc}
\begin{array}{l}
PE\left(\phi,\; 2i\right) = sin\left(sin\left(\phi\right) \cdot \alpha / 10000^{2i/d}\right) \\
PE\left(\phi,\; 2i+1\right) = sin\left(cos\left(\phi\right) \cdot \alpha / 10000^{2i/d}\right),
\end{array}
\end{equation}
\NEW{where $i \in [0,\; \frac{d}{2})$ and $d$ denotes the depth of the positional encoding. Note that $\alpha$ is an empirically determined parameter, which controls the width of the nonzero area of the encoding. The periodicity comes from the nested trigonometric function while uniqueness is established by interlacing the two functions. Fig.~\ref{fig:posenc} shows the positional encoding generated by the above function.}

\NEW{The resulting positional encoding of a subimage is the stacked vector of the three cyclic positional encodings. 
} 

\subsection{Calibration}
\NEW{Our neural network finally outputs a set of 3D coordinates which are the estimated sun directions of the given image patches in a sequence. Although this prediction was made by considering all patches from different time and space together, the estimates are in their own local camera coordinate systems. Therefore, we perform a calibration step using the camera ego-motion data to transform the estimated sun direction vectors into the world coordinate system.  
We assume the noise and drift in the ego-motion estimation is small relative to the lighting estimation. Hence, we employ a widely used structure-from-motion (SfM) technique such as \cite{schonberger2016structure} to estimate the ego-motion from an image sequence.
Each frames $f$ has a camera rotation matrix $R_f$ and the resulting calibrated vector $\hat{\vec v}_{pred}$ is computed as $R_f^{-1} \cdot \vec v_{pred}$.}

\NEW{Having the temporal estimates aligned in the same global coordinate system, we consider them as 
coherent observations of the same lighting condition in the temporal domain due to the spatio-temporal attention given from our transformer network. 
Finally, we take the mean of the individual aligned lighting estimates as our final prediction.}

\section{Experiments}
\label{sec:experiments}
\subsection{Datasets}
\NEW{We choose two datasets for evaluation: KITTI ~\cite{geiger2012we} and SUN360 ~\cite{reda2004solar}. KITTI is a popular dataset for autonomous driving. It consists of multiple driving sequences with rectified images and has additional annotations for determining the groundtruth sun directions ~\cite{reda2004solar}. This makes it an ideal candidate for testing our method on everyday driving scenes. For our experiments we create a random \textit{train-val-test} split composed of 47-5-5 driving scenes. This results in $\num{33889}$, $\num{3508}$, and $\num{3457}$ images respectively. Note that this \textit{scene} is different from the \textit{sequence} we give to the network. During the training and inference, we randomly select 8 frames from the same scene and generate 4 subimages from each image frame. Our pipeline estimates the global sun direction from this spatio-temporal sequence of 32 images.}


\NEW{SUN360~\cite{xiao2012recognizing} is one of the common datasets considered for outdoor lighting estimation methods. Several previous methods utilized it in its original panorama form or as subimages by generating synthetic perspective images~\cite{hold2017deep}. We followed the latter approach in our preliminary work~\cite{lee2021spatiotemporal} where we examined the performance improvement arising from the spatial aggregation. In particular, we simulate a camera motion without translation by generating a set of synthetic perspective images with a fixed field of view and randomized camera yaw and pitch angles. By doing so, we can perform the spatio-temporal aggregation on the SUN360 dataset in the same manner as on KITTI.
Specifically, we first divide $\num{12000}$ panorama images into the training, validation, and test sets with a 10:1:1 ratio. From each panorama, 16 perspective images with random yaw angles are taken where we match the horizontal and vertical field of views as the KITTI dataset. Note that we also introduce small random offsets on the camera elevation with respect to the horizon in [$-10^{\circ}$, $10^{\circ}$]. 
The generated images are resized to $1220 \times 370$ to match the size of the KITTI images. In this way, we produced $\num{160000}$, $\num{16000}$, and $\num{16000}$ images from $\num{10000}$, $\num{1000}$, and $\num{1000}$ panoramas for the training, validation, and test sets, respectively. The sequences for the training, validation, and test were generated in the same way as for the KITTI dataset. 
} 

\begin{figure}[!b]
	\begin{center}
		\centerline{\includegraphics[width=\textwidth]{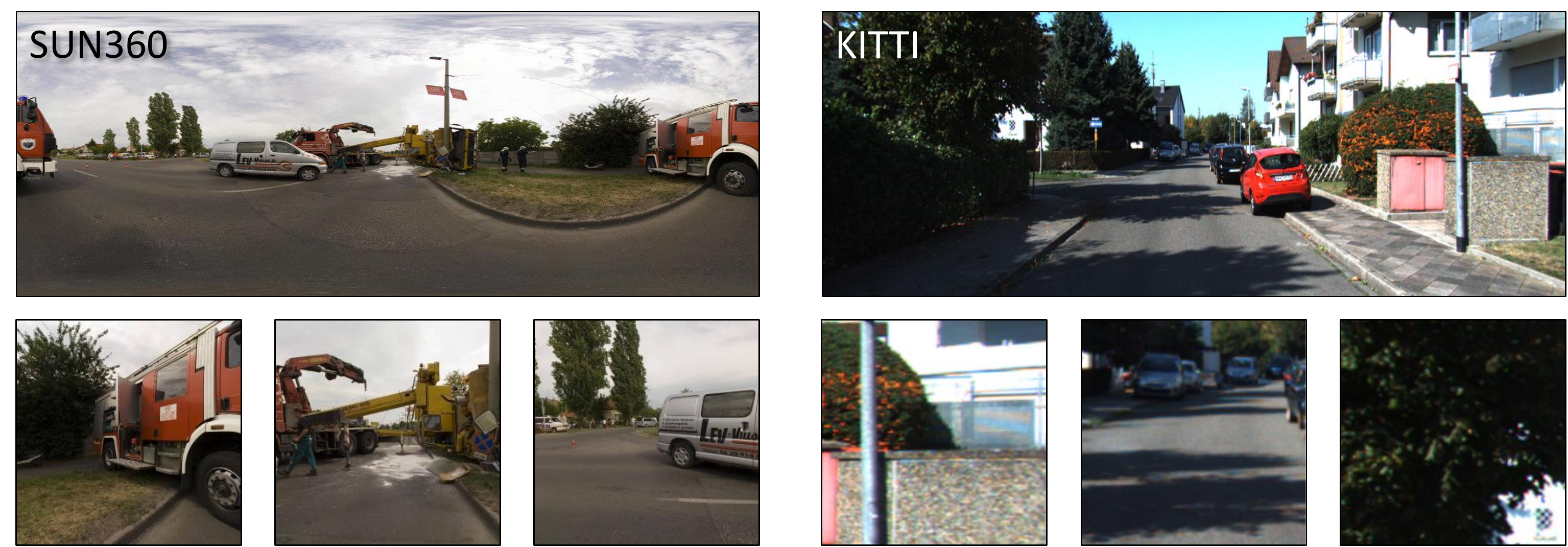}}
		\caption{Examples of the two datasets~\cite{geiger2012we,xiao2012recognizing}. From the original image (\emph{top}), we generate random subimages (\emph{bottom}).} 
		\label{fig:datasets}
	\end{center}
\end{figure}

\NEW{Since the volume of the SUN360 dataset is about 5 times larger than the KITTI dataset, we train and evaluate our network on the separated datasets. 
The exact numbers of panoramas and images are presented in Table \ref{tab:dataset} and 
Fig.~\ref{fig:datasets} illustrates examples from the two datasets.}

\begin{table}[tb]
	\begin{center}
		\centerline{\includegraphics[width=0.75\textwidth]{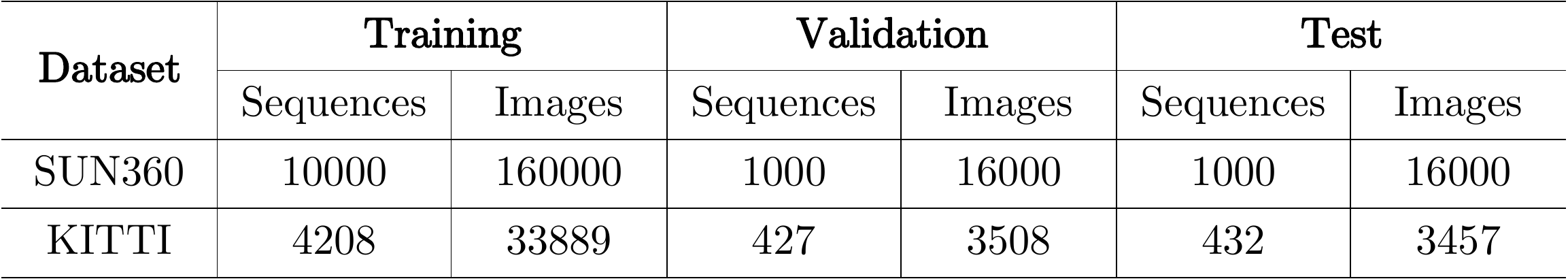}}
		\caption{Number of data in our datasets} 
		\label{tab:dataset}
	\end{center}
\end{table}

\subsection{Implementation Details}

\NEW{Our lighting estimation model consists of a ResNet18 encoder pretrained on the ImageNet dataset~\cite{deng2009imagenet} and a transformer network, followed by a dense layer converting a feature vector of dimension 512 to a 3D sun direction estimate. 
It accepts 32 RGB images of size $224 \times 224$ cropped from 8 frame images and outputs the sun direction estimate through the calibration and averaging process. We borrow the core structure of the transformer from \cite{dosovitskiy2020image} and carefully determine the number of layers, number of heads, hidden size, and MLP size as 4, 4, 512, and 1024, respectively, under extensive experiments. The dropout rate was 0.2. 
}


\NEW{We train our model and test its performance on the SUN360 and the KITTI datasets separately (see Table~\ref{tab:dataset}). In detail, we empirically trained our lighting estimation network for 13 and 47 epochs for the SUN360 and the KITTI datasets using early stopping. The training was initiated with the Adam optimizer~\cite{kingma2014adam} using a learning rate of $2\times10^{-5}$ and the batch size was 8. It took 15.2 and 6.4 hours on a single Nvidia RTX 3090 GPU. Prediction on a single sequence of 32 images takes 104\,ms. Our spatio-temporal aggregation model is examined upon \num{1000} unobserved SUN360 sequences and \num{432} KITTI sequences. 
} 

\subsection{Results}

\begin{figure}[bt!]
    \centering
	\includegraphics[width=\textwidth]{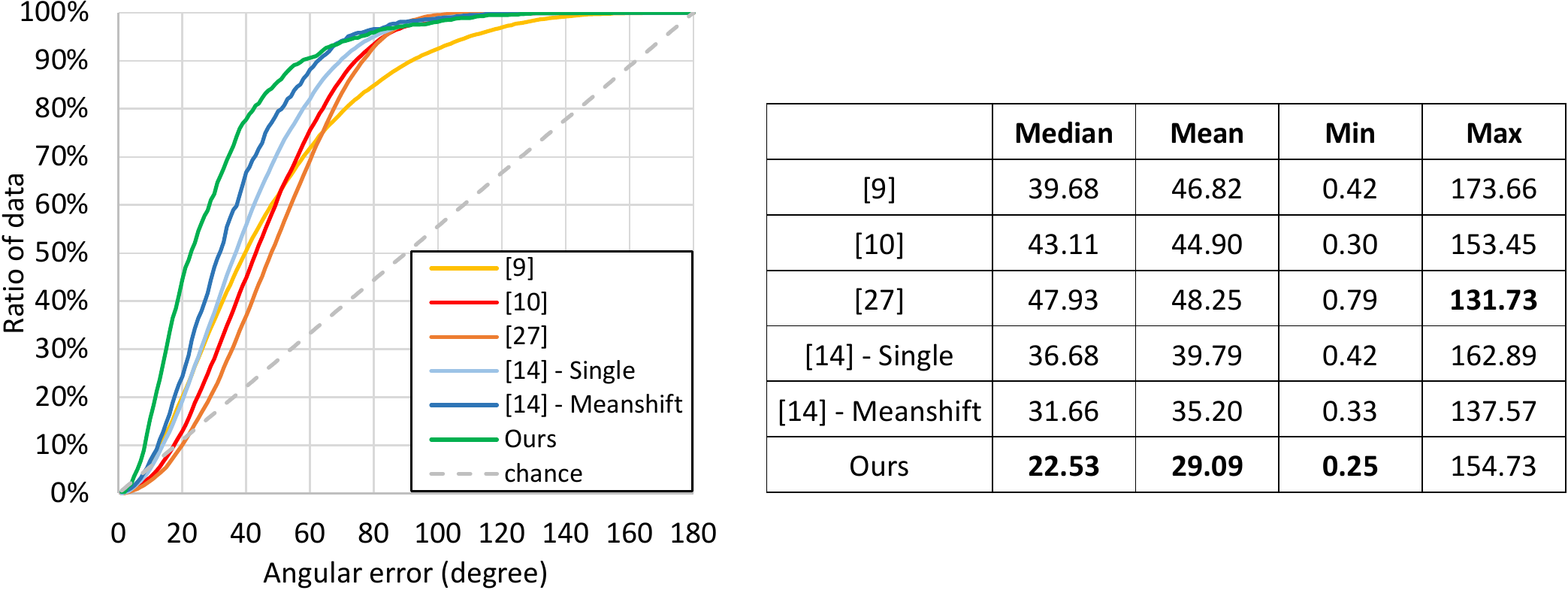} 
	\caption{\NEW{The cumulative angular error and the statistics of the sun direction estimates on the SUN360 test set. \textit{[14] - Meanshift} and \textit{Ours} are showing the spatiotemporal aggregation results. Note that performances of other methods are measured on single images although they can be boosted using the spatial aggregation method presented in \cite{lee2021spatiotemporal}. The proposed method outperforms our previous method with a noticeable margin.}} 
	\label{fig:sun_error}
\end{figure}

\NEW{We evaluate the angular errors of the spatio-temporally aggregated sun direction estimates on the SUN360 test sequences. At first, we present single image lighting estimation results from \cite{hold2017deep}, \cite{jin2020sun}, \cite{zhang2021outdoor}, and \cite{lee2021spatiotemporal}.
On top of that, we compare our method with the spatio-temporal aggregation pipeline proposed in \cite{lee2021spatiotemporal}. 
The hyperparameters required for our previous method are determined in the same way as decribed in \cite{lee2021spatiotemporal}.
} 

\NEW{Fig.~\ref{fig:sun_error} illustrates the cumulative angular errors of the five methods trained and tested on the SUN360 dataset. We present the outcomes of four single image based approaches including our previous work~\cite{lee2021spatiotemporal} along with the results of two spatio-temporal aggregation methods. Note that the performances of \cite{hold2017deep,zhang2021outdoor,jin2020sun} can potentially be improved by applying the statistical post-processing introduced in \cite{lee2021spatiotemporal}. Our spatio-temporal attention method shows a noticeable margin compared to the state-of-the-art.  
}




\begin{figure}[bt!]
    \centering
		\includegraphics[width=\textwidth]{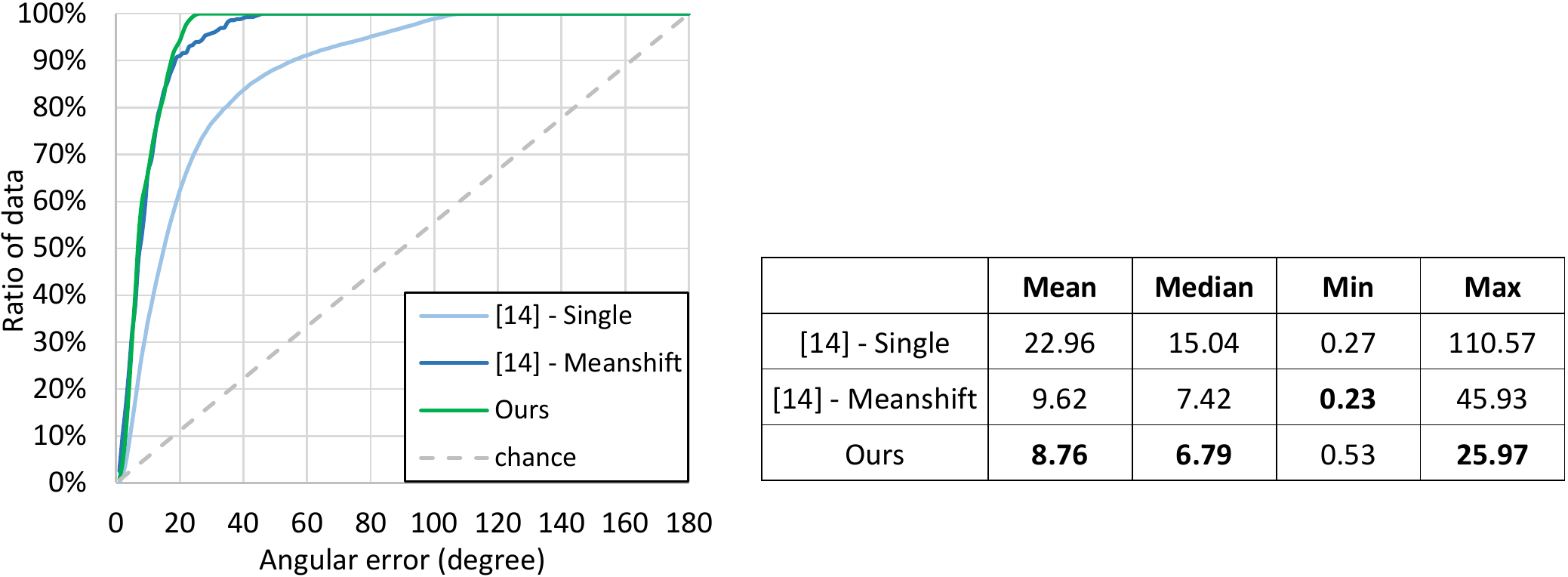}
		\caption{\NEW{The cumulative angular error and the statistics on the KITTI test set. Our method performs slightly better than \cite{lee2021spatiotemporal} while recording a noticeable small maximum angular error of $25.97^{\circ}$.}} 
		\label{fig:kitti}
\end{figure}

\NEW{We conduct a similar comparison also on the KITTI dataset (see Fig.~\ref{fig:kitti}). On this dataset, however, we compare our method only with \cite{lee2021spatiotemporal} due to the lack of ground truth information such as exposure and turbidity which are required for other previous works.
Although the dataset provides the ground truth ego-motion required for the calibration step, we calculated it using \cite{schonberger2016structure} to generalize our approach. The mean angular error of the estimated camera rotation using the default parameters was $1.01^{\circ}$ over the five test scenes.
Using the proposed spatio-temporal attention method, the mean angular error over the $432$ test sequences recorded $8.76^{\circ}$, which is marginally better than $9.62^{\circ}$ of \cite{lee2021spatiotemporal}.
}

\begin{figure}[bt!]
	\begin{center}
		\centerline{\includegraphics[width=\textwidth]{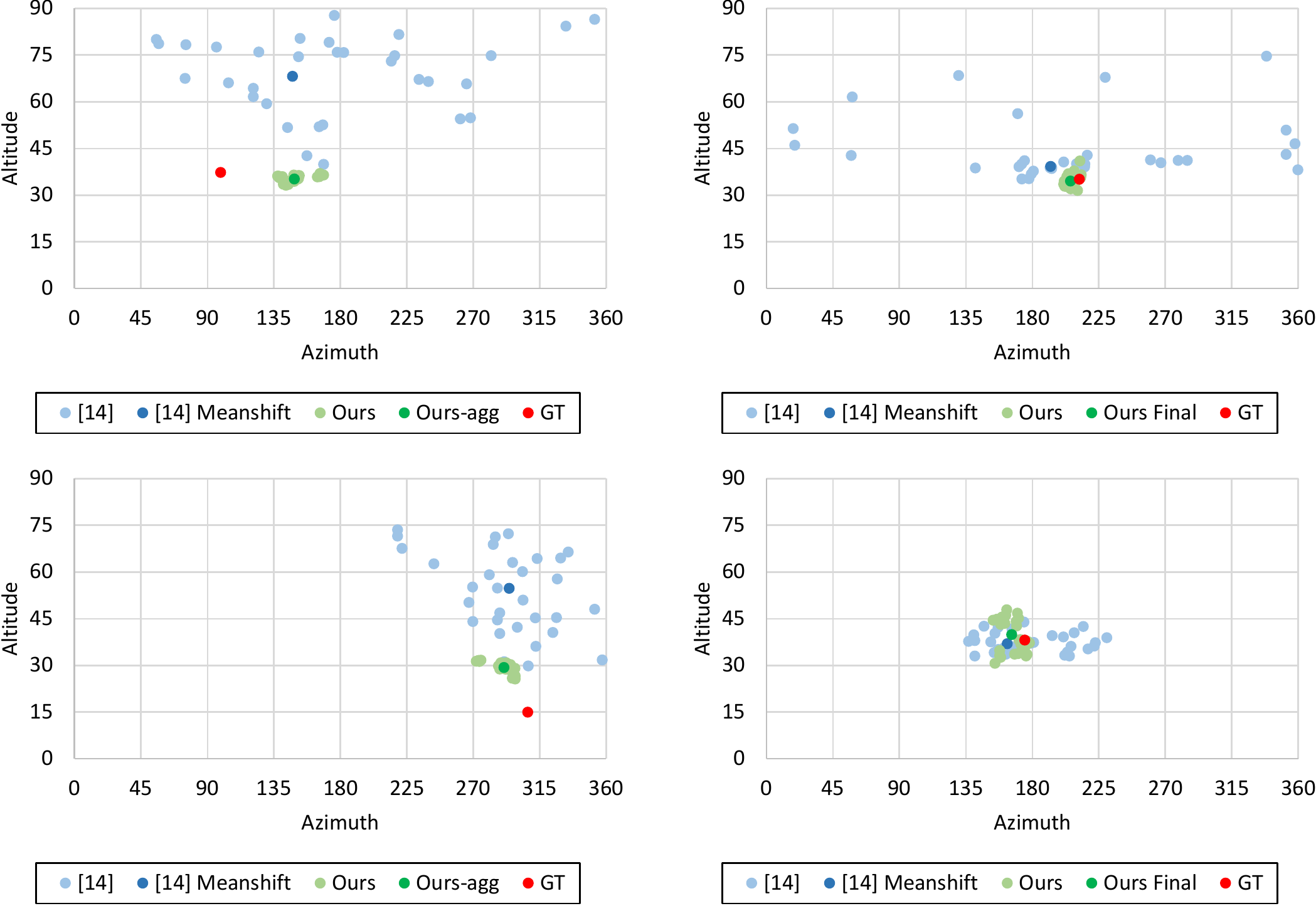}}
		\caption{Scatter plots representing sun direction estimates of individual subimages and the spatiotemporal aggregation result. Each plot corresponds to an image sequence of 8 frames in (\textit{left}) the SUN360 and (\textit{right}) the KITTI test sets. The spatio-temporal aggregation proposed in \cite{lee2021spatiotemporal} finds the highest point density among the inliers treating the estimates as independent sample. On the contrary, individual estimates of our method form a tight group due to the spatio-temporal attention. 
		} 
		\label{fig:calibration}
	\end{center}
\end{figure}

\NEW{We plotted the individual sun direction estimates and their aggregation results using our methods and \cite{lee2021spatiotemporal} in Fig.~\ref{fig:calibration}. Note that in the plots all predictions are registered to a common coordinate frame using the estimated camera ego-motion. Individual estimates of the subimages are shown with lighter color dots. The single image estimation of \cite{lee2021spatiotemporal} was done individually and resulted in independent noisy estimates which were aggregated using statistical post-processing. Unlike them, our estimates are jointly predicted and therefore tend to cluster tightly around their mean rendering any statistical post-processing redundant. This behavior comes from the spatio-temporal attention from our transformer network. We contend that the network tries to output a set of predictions that can explain the lighting condition of the given sequence, rather than predicting each subimage's lighting condition individually. Furthermore, this characteristic supports our decision for averaging all estimates to get the final estimate of the sequence. 
}

\begin{figure}[hbt!]
	\begin{center}
		\centerline{\includegraphics[width=0.85\textwidth]{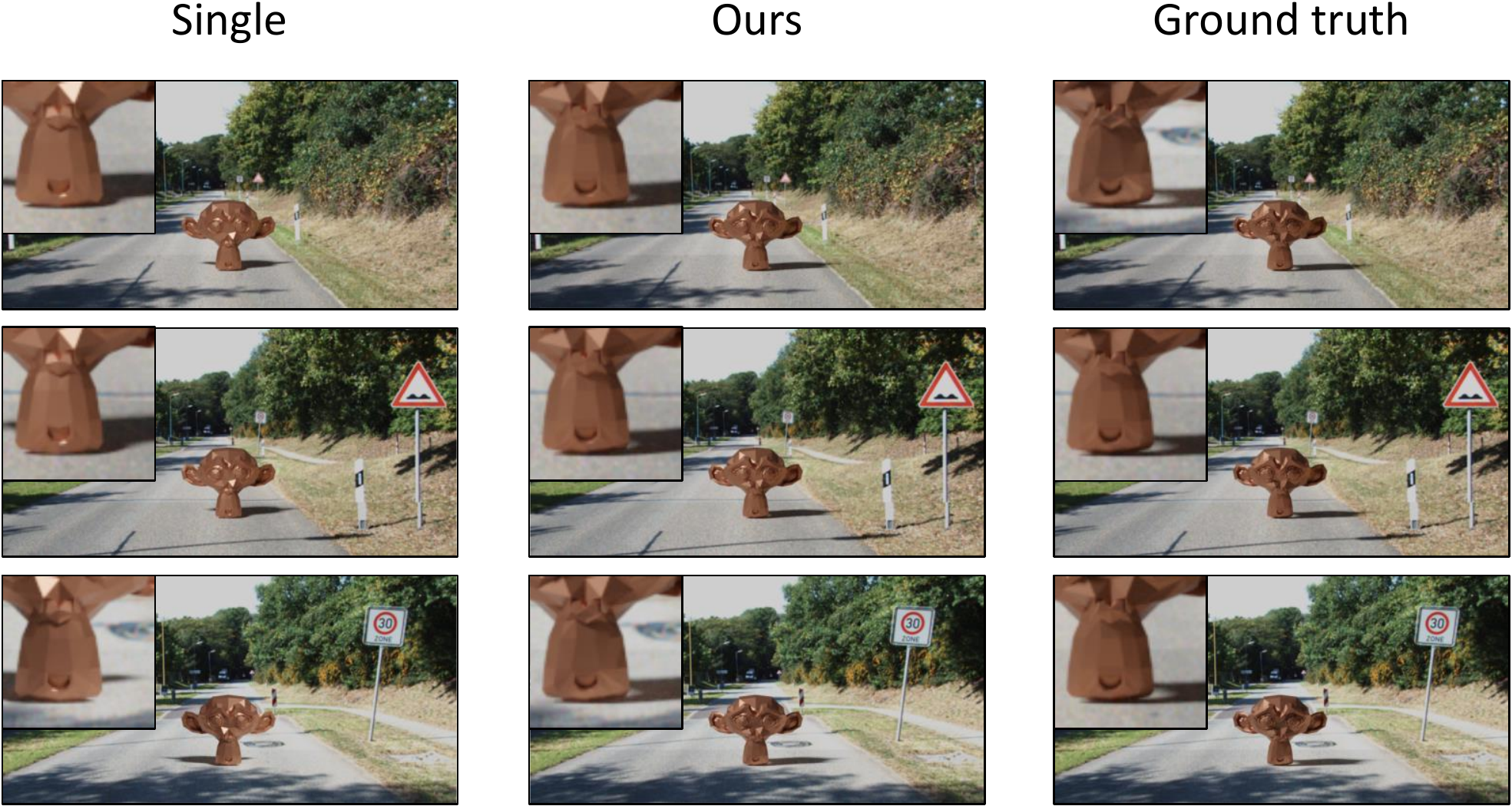}}
		\caption{\NEW{Demonstration of a virtual augmentation application. Fluctuations in the shadow of the augmented object are strongly visible when the sun direction estimates for individual images are applied. After applying our spatio-temporal aggregation, the results are fully stabilized and almost indistinguishable from the ground truth.}}
		\label{fig:augmentation}
	\end{center}
\end{figure}

	

Our model's stability is better understood with a virtual object augmentation application as shown in Fig.~\ref{fig:augmentation}. Note that other lighting parameters, such as the sun's intensity are manually determined. When the lighting conditions are estimated from only a single image on each frame, the virtual objects' shadows are fluctuating compared to the ground truth results. The artifact is almost entirely removed after applying the spatio-temporally aggregated lighting condition.

\section{Conclusion}
\NEW{In this paper, we proposed a holistic sequence-wise lighting estimation method based on spatio-temporal attention using transformers. Our method achieved state-of-the-art performance on outdoor lighting estimation for a given image sequence. 
Without loss of generality we utilized $360^{\circ}$ panoramas and wide view images in our work, but the method can also be applied to any image providing enough details. Moreover, our spatio-temporal aggregation could also be generalized to other globally shared image information under given computational budgets.}

Although we demonstrated noticeable outcomes in augmented reality applications, intriguing future research topics are remaining. 
We plan to extend our model to examine other factors such as cloudiness or exposure as it helps to accomplish diverse targets, including photorealistic virtual object augmentation across an image sequence. With such augmented datasets, we could enhance the performance of other deep learning techniques.
And last, knowing the lighting in the 3D scene behind an image can facilitate shadow detection or removal algorithms and help initializing global camera orientation estimation in SLAM approaches. 


\bibliographystyle{unsrt}  

\bibliography{references}

\end{document}